\documentclass{article}

\usepackage{arxiv}

\usepackage{framed,multirow}
\usepackage{amssymb}
\usepackage{latexsym}
\usepackage{amsfonts}
\usepackage{amsmath,amsfonts}
\usepackage{algorithmic}
\usepackage{array}
\usepackage[caption=false,font=normalsize,labelfont=sf,textfont=sf]{subfig}
\usepackage{textcomp}
\usepackage{stfloats}
\usepackage{url}
\usepackage{verbatim}
\usepackage{graphicx}
\usepackage{xcolor}
\usepackage{svg}
\usepackage{calc}
\usepackage{multicol}
\usepackage{accents}
\usepackage{footnote}
\usepackage{amstext}
\usepackage{stmaryrd}
\usepackage[mathscr]{euscript}
\usepackage{bigints}
\usepackage{upgreek}
\usepackage[percent]{overpic}
\usepackage{soulutf8}
\usepackage{bm}
\usepackage{physics}
\usepackage{mathtools}
\usepackage{caption}
\usepackage{yhmath}
\usepackage{tikz}
\usepackage{comment}
\usepackage{cellspace}
\usepackage{hyperref}
\usepackage{tabularx}
\usepackage{adjustbox}
\usepackage{colortbl}
\usepackage{mwe}
\usepackage[utf8]{inputenc}
\usepackage[T1]{fontenc}
\usepackage{hyperref}
\usepackage{url}
\usepackage{booktabs}
\usepackage{amsfonts}
\usepackage{nicefrac}
\usepackage{microtype}
\usepackage{cleveref}
\usepackage{lipsum}
\usepackage{natbib}
\usepackage{doi}

\setcitestyle{numbers,square}

\title{RoCNet: 3D robust registration of point clouds using deep learning}

\newif\ifuniqueAffiliation
\uniqueAffiliationtrue

\ifuniqueAffiliation 
\author{ {\textbf{Karim Slimani}}\thanks{ authors are with Sorbonne Universit\'e, CNRS UMR 7222, INSERM U1150, ISIR, F-75005, Paris, France} \\
	ISIR, Sorbonne Université,\\
	  CNRS UMR 7222, INSERM U1150\\
	4 place Jussieu , Paris , 75005 \\
	\texttt{karim.slimani@isir.upmc.fr} \\
	\And
	 {\textbf{Catherine Achard}} \\
	ISIR, Sorbonne Université,\\
	  CNRS UMR 7222, INSERM U1150\\
	4 place Jussieu , Paris , 75005 \\
	\texttt{catherine.achard@sorbonne-universite.fr} \\
	\And
	{\hspace{1mm}Brahim Tamadazte} \\
        ISIR, Sorbonne Université,\\
	  CNRS UMR 7222, INSERM U1150\\
	4 place Jussieu , Paris , 75005 \\
	\texttt{{{brahim.tamadazte@cnrs.fr}}} \\
}
\else
\usepackage{authblk}

\newbox{\orcid}\sbox{\orcid}{\includegraphics[scale=0.06]{orcid.pdf}} 
\author[1]{%
	\href{https://orcid.org/0000-0000-0000-0000}{\usebox{\orcid}\hspace{1mm}David S.~Hippocampus\thanks{\texttt{hippo@cs.cranberry-lemon.edu}}}%
}
\author[1,2]{%
	\href{https://orcid.org/0000-0000-0000-0000}{\usebox{\orcid}\hspace{1mm}Elias D.~Striatum\thanks{\texttt{stariate@ee.mount-sheikh.edu}}}%
}
\affil[1]{Department of Computer Science, Cranberry-Lemon University, Pittsburgh, PA 15213}
\affil[2]{Department of Electrical Engineering, Mount-Sheikh University, Santa Narimana, Levand}
\fi

\hypersetup{
pdftitle={A template for the arxiv style},
pdfsubject={q-bio.NC, q-bio.QM},
pdfauthor={David S.~Hippocampus, Elias D.~Striatum},
pdfkeywords={First keyword, Second keyword, More},
}

\begin{document}
\maketitle

\begin{abstract}
This paper introduces a new method for 3D point cloud registration based on deep learning. The architecture is composed of three distinct blocs: (\emph{i}) an encoder with a convolutional graph-based descriptor that encodes the immediate neighbourhood of each point and an attention mechanism that encodes the variations of the surface normals. Such descriptors are refined by highlighting attention between the points of the same set (source and target) and then between the points of the two sets. (\emph{ii}) a matching process that estimates a matrix of correspondences using the Sinkhorn algorithm. (\emph{iii}) Finally, the rigid transformation between the two point clouds is calculated by RANSAC using the best scores of the correspondence matrix. We conduct experiments on the ModelNet40 dataset and real world Bunny dataset, and our proposed architecture shows very promising results, outperforming state-of-the-art methods in most of the simulated configurations, including partial overlap and data augmentation with Gaussian noise.
\end{abstract}

\keywords{Point Cloud Registration \and  Deep Learning \and  Attention Mechanism \and  Pose Estimation}

\section{Introduction}

{P}{oint} cloud registration is a  widespread problem and a key task in robotics and computer vision, with applications in pose estimation, 3D reconstruction, 3D localisation etc. The registration process involves matching points between the source and target point clouds, eliminating outliers, and estimating the rigid transformation parameters that aligns one point cloud to the other. Traditional algorithms, such as Iterative Closest Point (ICP)~\citep{besl1992method}, alternate between matching and aligning iterations. Recently, neural network-based techniques, such as~\citep{wang2019prnet, Wang_2019_ICCV, aoki2019pointnetlk}, are increasingly used to address these problems. These techniques typically encode each point and its neighbourhood through a learned descriptor, such as~\citep{dgcnn, qi2017pointnet++}.
Point matching is often performed based on similarities in the features space.
The estimation of the rigid transformation can be achieved in two different ways, either using end-to-end learning by integrating differentiable SVD~\citep{soft_svd} into the network as proposed in~\citep{Wang_2019_ICCV}, or by applying a simple SVD or a RANSAC.

The learning-based approaches have made significant progresses and have led to overcome numerous limitations of iterative methods, such as converging to a local minimum, conditioned by a correct initialization. Additionally, most of iterative methods extract point-wise features by applying hand-crafted or learned descriptors. This makes these methods more sensitive to noise and outliers. The methods that apply RANSAC to overcome this problem  need a huge number of iterations (e.g., 50, 000) to reach a correct estimation of the transformation. 

\begin{figure*}
\centerline{\includegraphics[width=1\columnwidth]{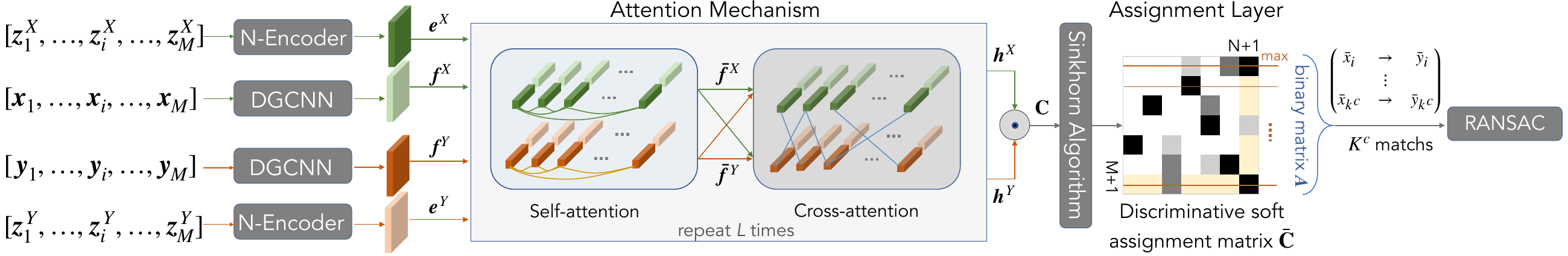}}
\caption{Overview of the proposed RoCNet architecture.}
\label{fig.architecture}
\end{figure*}

In this paper, we propose a new architecture (Fig.~\ref{fig.architecture}), called RoCNet, which includes three main blocks: 1) a descriptor composed of a convolutional graph-based network that encodes the immediate neighbourhood of each point and an attention mechanism that encodes the variations of the surface normals, 2) a matching module that estimates a matrix of correspondences using the Sinkhorn algorithm, 3) a RANSAC module which computes the rigid transformation using the $K^c$ (e.g., 256) best matches with a limited number of iterations (e.g., 500 iterations). The proposed architecture was assessed using ModelNet40 dataset~\citep{wang2019prnet} in different favourable and unfavourable conditions and Stanford Bunny dataset~\citep{turk1994zippered}. Our method outperforms the related state-of-the-art algorithms, especially in unfavourable conditions, e.g., with noisy data and partial occlusions.
%

\section{RELATED WORK}\label{sec.related}
%
This section provides a review of the state-of-the-art regarding the main methods for 3D pose estimation and point cloud registration. 

%
\subsection{3D Point Cloud Descriptors} 

{The first iterative methods presented below, use the 3D coordinates points directly as input to their system or handcrafted features like Fast Point Feature Histograms (FPFH)~\citep{rusu2009fast}. Since the rise of deep learning (DL), several methods have been developed to learn 3D point cloud descriptors. 
For instance, PointNet~\citep{qi2016pointnet} or its extension PointNet++~\citep{qi2017pointnet++} describes an unordered 3D points set for tasks such as classification, segmentation, and registration. 
Wang \textit{et al.}~\citep{dgcnn} propose the Dynamic Graph CNN (DGCNN) descriptor based on a module that acts on graphs capturing semantic characteristics over potentially long distances.
In~\citep{mdgat},  authors build a feature with two main components, a descriptor encoder that highlights handcrafted features obtained with FPFH and a positional encoder that highlights spatial properties of the point cloud.}

%
\subsection{Iterative Methods}\label{Iterative Methods}

The ICP~\citep{besl1992method} is probably the most popular method to address a point cloud registration problem. Given two sets of 3D points, the purpose is to minimise the Euclidean distance between the points. At each iteration, a mapping of the two sets of points and the computation of the 3D rigid transformation using an SVD are performed. This procedure is repeated until convergence. In RANSAC, the two-point clouds are randomly split into subsets on which a transformation is estimated. The final transformation is chosen among them using a {predefined criterion such as the number of inliers provided by each transformation.} Both methods have been associated with neural network-based learned descriptors like in~\citep{Wang_2019_ICCV,r_pointhop} for ICP or in~\citep{li2022wsdesc} for RANSAC. 

%
\subsection{Methods based on Matching Learning} 
Recent registration methods investigate DL architecture to match the points. For instance, Predator~\citep{huang2021predator} is trained with three different weighted matching losses to be more robust to low partial overlap. 
Alternatively, in 3DFeat-Net~\citep{yew2018-3dfeatnet}, the architecture is trained to detect key points and predict discriminative descriptors using the triplet loss~\citep{triplet_loss}, while works in~\citep{bai2020d3feat} combine two losses, one for the descriptor and the other one for the detector. All these methods use RANSAC on matched features  to estimate the transformation parameters. MDGAT~\citep{mdgat} learn the matching using a new loss inspired by the triplet function.~\citep{roufosse2019unsupervised} propose an unsupervised matching approach by optimizing the structural properties (i.e  bijectivity or approximate isometry) of functional maps~\citep{ ovsjanikov2012functional}. In GeoTransformer~\citep{geotransformer}, super-points are extracted from the source point clouds and described using the Geometric Transformer which encodes intra-point-cloud and inter-point-cloud geometric structures. 
%
\subsection{Methods based on Transformation Learning} 
These methods are end-to-end and learn directly the transformation. For example, Deep Closest Point (DCP)~\citep{Wang_2019_ICCV} uses a descriptor based on DGCNN and a soft SVD. In PRNET~\citep{wang2019prnet}, the matching is performed using an approximately differentiable version of Gumbel-Softmax and the transformation is also obtained using a SVD. Another approach, proposed in~\citep{li2022wsdesc}, uses a differentiable nearest neighbour search algorithm in the descriptor space to match the points, and then proposes to relax the registration problem and seeks to estimate an affine transformation matrix computed by a 
least squares optimisation. SCANet~\citep{zhou2021scanet} introduces a new channel cross-attention regression (CCR) module for pose estimation allowing the model to aggregate cross point clouds information and then predicting the rotation quaternions and the translation vector in a fully differentiable manner.

\section{Method}\label{sec.method}

\subsection{Problem Statement}\label{sub.sec.problem}
Let us start by defining a common problem of 3D point cloud registration. Considering two-point clouds $\boldsymbol{X}$ and $\boldsymbol{Y}$ such that: 
$\boldsymbol{X}=\{ \boldsymbol{x}_1,...,\boldsymbol{x}_i,...,\boldsymbol{x}_M\} \subset \mathbb{R}^{3\times M}$ and $ \boldsymbol{Y}=\{ \boldsymbol{y}_1,...,\boldsymbol{y}_j,...,\boldsymbol{y}_N\} \subset \mathbb{R}^{3\times N}$. It is assumed that the two sets at least partially overlap, so that there are $K$ pairs of matches between $\boldsymbol{X}$ and $\boldsymbol{Y}$, with $K \leq min(M,N)$. The two subsets containing the matching points in the first and second point clouds are defined by: $\Bar{\boldsymbol{X}} \subset \mathbb{R}^{3\times K} $ and $ \Bar{\boldsymbol{Y}} \subset \mathbb{R}^{3\times K}$, respectively. Note that the set $\Bar{\boldsymbol{Y}}$ is obtained by applying a rotation $\bold{R} \in SO(3)$ and a translation $\bold{t} \in \mathbb{R}^{3}$ of the set $\Bar{\boldsymbol{X}}$. Both the rotation matrix $\bold{R}$ and translation vector $\bold{t}$ define the $4 \times 4$ rigid transformation we are looking for. 
%
%
\subsection{Descriptor}\label{subsec.descriptor}
%
One of the most fundamental components in a point cloud registration problem is the relevance and quality of the descriptor used to encode the points. Therefore, we proposed a new descriptor by projecting the source and target sets of points $\boldsymbol{X}$ and $\boldsymbol{Y}$ in a new base of higher dimension, de facto, more discriminating than the initial spatial representation, and as invariant as possible to rotations and translations. It combines a geometrical-based descriptor and a normal-based one, followed by an attention mechanism.
%
\subsubsection{Geometrical-based descriptor}\label{DGCNN} 
Different types of descriptors, that learn local geometrical properties around each point, were reported in the literature such as PointNet~\citep{qi2016pointnet}, PointNet++~\citep{qi2017pointnet++} or DGCNN descriptor~\citep{dgcnn}.
We integrate DGCNN as a part of our descriptor because it better captures local geometric features of point clouds while still maintaining permutation invariance. It consists of mainly \emph{EdgeConv} convolution layers where the points represent nodes connected by arcs to their $k$ nearest neighbours in the encoding space to build graphs that express the local geometric structure surrounding each point and then spread dynamically the information at a higher level (global encoding). Let us denote $\boldsymbol{f}^X_i$ the extracted feature vector of dimension $d$ for point $\boldsymbol{x}_i$.
\subsubsection{Normal based descriptor}\label{sub.sec.normals}
The main idea of this descriptor is to better encode the surface around each point using the variation of the normals of points in the neighbourhood: in a flat surface, there is no variation of the normals, along a ridge, the normals vary only in one direction, whereas on a summit, the normals vary in all directions. Thus, the variation of the angle of the normals in a neighbourhood is informative about the type of surface.

The normals are estimated using Principal Component Analysis (PCA). Indeed, for each point $\boldsymbol{x}_i \in \boldsymbol{X}$, a local neighbourhood subset of points $S_i = \left \{ \boldsymbol{x}_j /  \left\| \boldsymbol{x}_j - \boldsymbol{x}_i\right\| ^2 \leq r \right \}$ is defined while delimiting the size of the set points with $|S_i|<K_{nn}$. $r$ is the radius of a sphere centred on $\boldsymbol{x}_i$ and $K_{nn}$ the maximum number of points included in the set $S_i$. The eigenvalue decomposition of the covariance matrix $Cov(S_i)$ allows defining the normal $\boldsymbol{n}_i$ as the vector associated with the smallest eigenvalue. 

Since the PCA does not inherently determine the direction of the normal vector, we propose to address the ambiguity of the sign by using a new vector $\boldsymbol{z}_i$. It is colinear to $\boldsymbol{n}_i$ and is defined by ensuring that it points towards the side of the surface with a higher point density. This means that the normal vector points away from sparse areas and towards denser surface areas. Similar to~\citep{li2022wsdesc}, we solve this ambiguity thanks to the system:  
\begin{equation}
\boldsymbol{z}_i = 
\begin{cases}
\boldsymbol{n}_i, & \text{if } ~~\underset{{\boldsymbol{x}_j \in S_i}}\sum \boldsymbol{n}_i^T ~( \boldsymbol{x}_i - \boldsymbol{x}_j ) \geq 0 \\\
-\boldsymbol{n}_i, & \text{otherwise}
\end{cases}
\end{equation}
  
Finally, we build the final encoding based on~\citep{geotransformer} and~\citep{vaswani2017attention} using sinusoidal functions of different frequencies. Knowing the angle between the normals of two points $\boldsymbol{x}_i$ and $\boldsymbol{x}_j$ noted $\angle(\boldsymbol{z}_i,\boldsymbol{z}_j)$, the vector $\boldsymbol{g}_{\boldsymbol{x}_i,\boldsymbol{x}_j}$ encoding the normals is given by:  
\begin{equation} \label{eq.normal_embed}
\begin{cases}
\boldsymbol{g}_{\boldsymbol{x}_i,\boldsymbol{x}_j}^{2 ind} = \sin\left( \frac{\angle(\boldsymbol{z}_i,\boldsymbol{z}_j) } {\tau \times 10000^{2 ind/{d}} } \right) \\
\boldsymbol{g}_{\boldsymbol{x}_i,\boldsymbol{x}_j}^{2 ind+1} = \cos \left( \frac{ \angle(\boldsymbol{z}_i,\boldsymbol{z}_j)} {\tau \times 10000^{2 ind/{d}} } \right)
\end{cases}   
\end{equation}

where $ind$ is the current value index of $\boldsymbol{g}_{\boldsymbol{x}_i,\boldsymbol{x}_j}$, $\tau$ a normalisation coefficient and $d$ the dimension of the descriptor $\boldsymbol{g}_{\boldsymbol{x}_i,\boldsymbol{x}_j}$ fixed to the same as the DGCNN output. A fully connected layer is then applied to $\boldsymbol{g}_{\boldsymbol{x}_i,\boldsymbol{x}_j}$ to obtain the final embedding 
$    \boldsymbol{e}_{i,j}^X = \boldsymbol{g}_{\boldsymbol{x}_i,\boldsymbol{x}_j} \bold{W}^s_E $
where $\bold{W}^s_E \in \mathbb{R}^{d \times d}$ is a learned projection matrix.
%
\subsubsection{Attention mechanism} \label{normal_transf}
A key point of recent descriptors of point cloud is the introduction of an attention mechanism that highlights some features dynamically.
SuperGlue~\citep{sarlin20superglue} uses a module based on attention graphs that alternately stacks 'self-attention' and 'cross-attention' layers. The former links all the nodes of a point cloud to each other, while the latter links each point of set $\boldsymbol{X}$  to all points of set $\boldsymbol{Y}$. 
Contrary to SuperGlue~\citep{sarlin20superglue} or MDGAT~\citep{mdgat}, which compute the attention weights on the encoding vectors, some methods propose adding information on local inter-points geometry at the entry of the mechanism.  
For instance, \citep{zhao2021point} associates the 3D coordinates of each point with the descriptor, while GeoTransformer~\citep{geotransformer} proposes to use the distances and angles between each point and its $k$ nearest neighbours. Alternatively, in our approach, we propose the use of four attention heads with geometric self-attention inside each set $\boldsymbol{X}$ and $\boldsymbol{Y}$ integrating the associated normals embeddings $\boldsymbol{e}^X$ and $\boldsymbol{e}^Y$ respectively, followed by a cross-attention between the two sets of points and then alternate between them for $L$ times.
\subsubsection{Self-attention} 
This type of layer predicts an attention-based feature $\bar{\boldsymbol{f}}_i$ for each point of a point cloud ($\boldsymbol{X}$ or $\boldsymbol{Y}$), paying attention to all the other points of the same cloud. In the following, the algorithm is detailed for a point $\boldsymbol{x}_i \in \boldsymbol{X}$, the same is used for all the points in $\boldsymbol{X}$  and $\boldsymbol{Y}$. Thus, an attention weight is  obtained for each query/key pair:
\begin{equation}\label{wei}
\alpha_{ij}^X =  \underset{j}{softmax} \bigg( \frac{(\boldsymbol{f}_i^X \bold{W}^s_Q )(\boldsymbol{f}_j^X \bold{W}^s_K+\boldsymbol{e}_{i,j}^X \bold{W}^s_R)^T}{\sqrt{{d}}}\bigg)
\end{equation}
where $\bold{W}^s_Q$, $\bold{W}^s_K$ and $\bold{W}^s_R \in \mathbb{R}^{d\times d}$ are the learned projection matrices for queries, keys and normal-based embeddings, $d$ is the dimension of the features $\boldsymbol{f}_i^X$ and $\boldsymbol{e}^X_{i,j}$. These weights are used to rate which elements we have to pay attention to, and to obtain the final self-attention-based feature $\bar{\boldsymbol{f}}_i^X$: 
\begin{equation}\label{wei}
\bar{\boldsymbol{f}}_i^X =   \sum_{j=1} \alpha_{ij} \boldsymbol{v}_j
\text{ with, } 
\boldsymbol{v}_j = \boldsymbol{f}_j^X \bold{W}^s_V
\end{equation}
where $\bold{W}^s_V \in \mathbb{R}^{d\times d}$  is the learned projection matrix for values. 
%
\subsubsection{Cross-Attention} 
A \emph{cross-attention} layer is used to propagate the local information between the two previously obtained representations $\bar{\boldsymbol{f}}_i^X$ and $\bar{\boldsymbol{f}}_j^Y$ of $\boldsymbol{x}_i$ and $\boldsymbol{y}_j$ belonging respectively to point clouds $\boldsymbol{X}$ and $\boldsymbol{Y}$. Formally, it works similarly as the \emph{self-attention} layer, except for the estimation of the attention key, which now uses a point in the second point cloud. The final encoding for any point $\boldsymbol{x}_i$ (or $\boldsymbol{y}_j)$ is given by:

\small
\begin{equation}\label{wei}
{\boldsymbol{h}}_i^X =   \sum_{j=1}^{\lvert  \boldsymbol{Y} \rvert } \Bigg( \underset{j}{softmax} \bigg( \frac{(\bar{\boldsymbol{f}}_i^X \bold{W}^c_Q )(\bar{\boldsymbol{f}}_j^Y \bold{W}^c_K)^T}{\sqrt{{d}}}\bigg) \Bigg ) \Bigg ( \bar{\boldsymbol{f}}_j^Y \bold{W}^c_V \Bigg )
\end{equation}
\normalsize
where $\bold{W}^c_Q$, $\bold{W}^c_K$ and $\bold{W}^c_V \in \mathbb{R}^{d\times d}$ are the learned projection matrices for queries, keys and values in the cross-attention layers.
%
\subsection{Point Matching} \label{matching}
The second step of the proposed algorithm is the matching procedure. We first estimate a score matrix  $\bold{C} \in \mathbb{R}^{M\times N}$ between each point $x_i \in \boldsymbol{X}$ and $y_j \in \boldsymbol{Y}$: 
\begin{center}
\begin{equation}\label{eq.score_matrix}
\bold{C}_{i,j} = {\boldsymbol{h}_i^X}^\top \boldsymbol{h}_j^Y
\end{equation}
\end{center}
where $\boldsymbol{h}_i^X$ and $\boldsymbol{h}_j^Y$ are the final encoding of the points $\boldsymbol{x}_i$ and $\boldsymbol{y}_j$ defined previously. To build a matrix of correspondence probabilities $\bold{\Bar{C}}$, we first augment the dimensions of $\bold{C}$ to $M+1$ and $N+1$ respectively, such that the non-matched points will explicitly be assigned to the last dimension. We then employ the differentiable \emph{Sinkhorn Algorithm}~\citep{sinkhorn1967concerning} which is widely used in optimal transport and graph-matching problems.

As all the previous steps are differentiable, the weights of the networks can be learned by introducing a loss function. To do so, we follow~\citep{mdgat} and adopt the gap loss function~(\ref{gap_loss_equation}) which allows enlarging the assignment scores difference between the true matches and the wrong matches. It is expressed as follows:

\small
\begin{multline}\label{gap_loss_equation}
    L_{Gap} = 
    \sum_{i=1}^M \log ( \sum_{n=1}^{N+1}[\max((-\log \bold{\Bar{C}}_{i,\bar{i}} + \log \bold{\Bar{C}}_{i,n} +\alpha),0)]+1 ) \\ + 
    \sum_{j=1}^N \log ( \sum_{n=1}^{M+1}[\max((-\log \bold{\Bar{C}}_{j,\bar{j}} + \log \bold{\Bar{C}}_{n,j} +\alpha),0)]+1 ) 
\end{multline}
\normalsize
where $\alpha$ is a positive scalar having a value of 0.5, $\bold{\Bar{C}}_{i,\bar{i}}$ and $\bold{\Bar{C}}_{j,\bar{j}}$ are the scores for the ground truth true matches of the points $\boldsymbol{x}_i$ and $\boldsymbol{y}_j$, respectively.

\begin{figure}[!h]
\centering
\begin{tabular}{ccc}
\includegraphics[width=0.2\columnwidth]{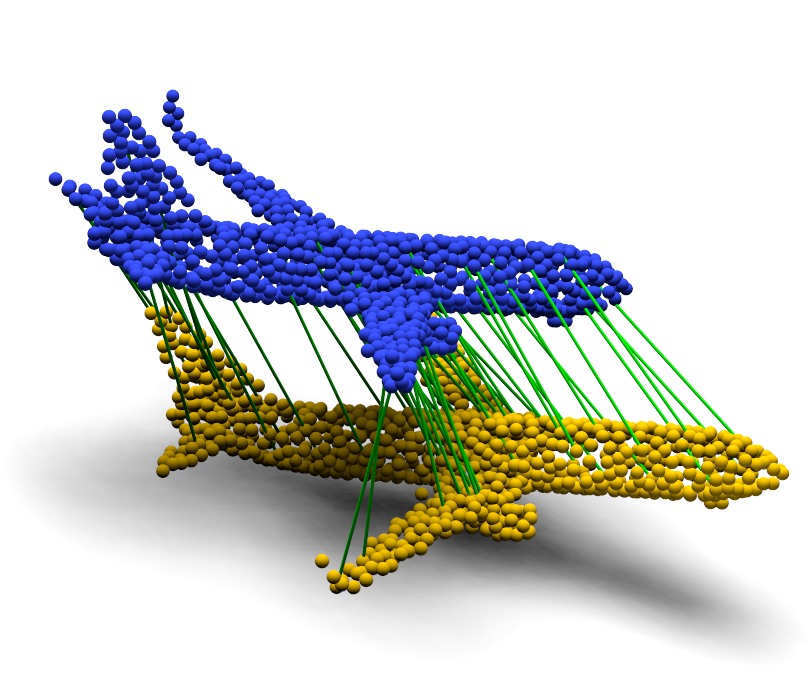} &
\includegraphics[width=0.17\columnwidth]{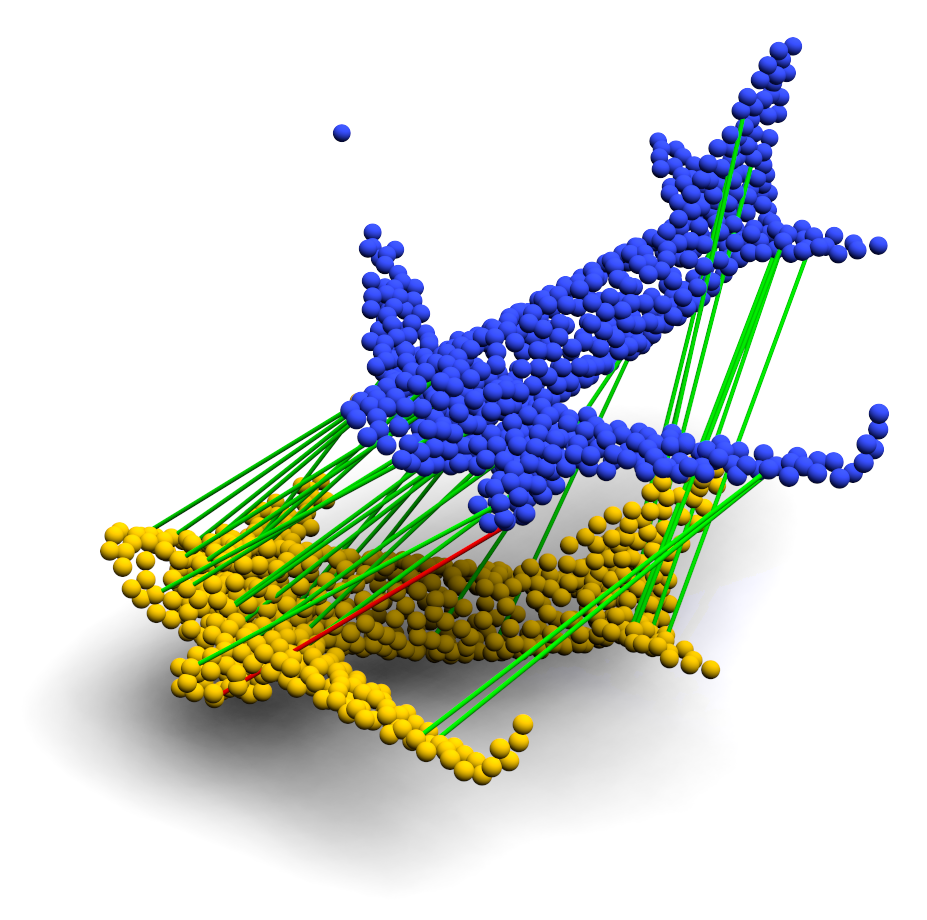} &
\includegraphics[width=0.17\columnwidth]{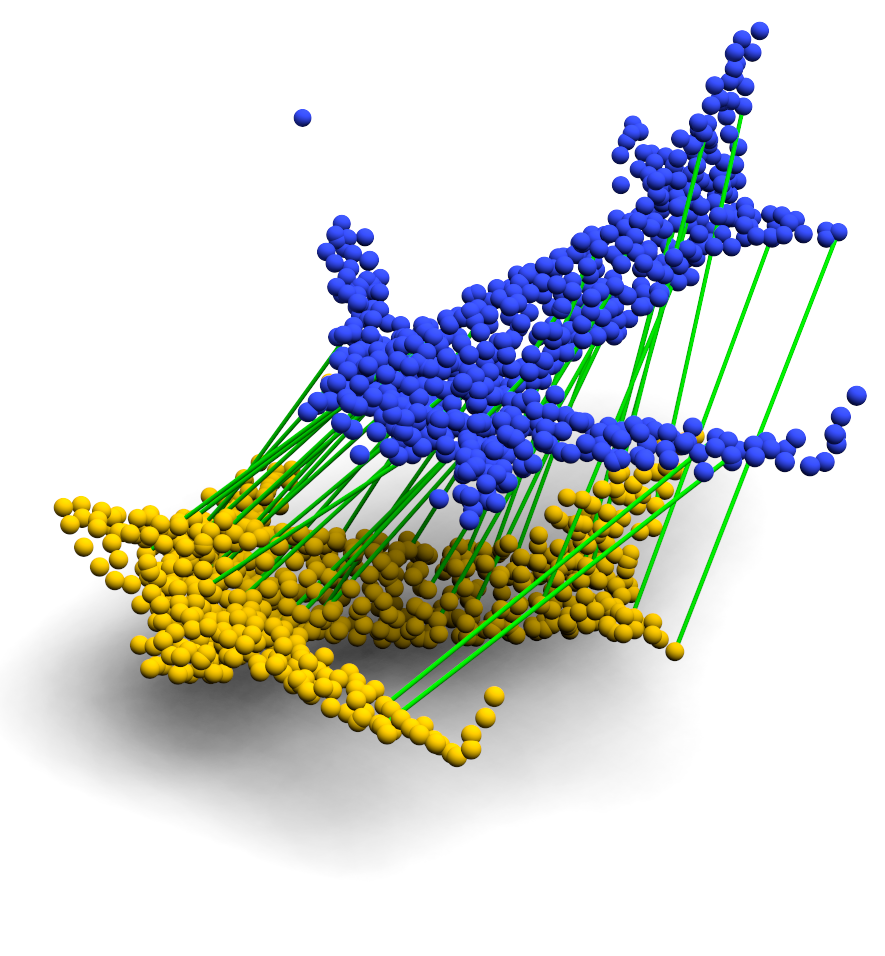} \\[-2ex] 
{(a)} & {(b)} & {(c)} \\
\end{tabular}
\caption{Example of a performed 3D matching of point clouds in different configurations: (a) clean data, (b) partial overlap, and (c) noisy data and partial overlap. The green lines show the correct matches and the red lines show the wrong ones. }
\label{fig.matching}
\end{figure}
%
 \subsection{Pose Estimation}\label{RANSAC}

In the evaluation phase, we build a hard assignment binary matrix $\bold{A}$ thanks to the following algorithm: 
{\small
\begin{equation}
\begin{aligned}
    \bold{A^{1}}_{i,j} &= \begin{cases} 1 &\text{if } \bold{\Bar{C}}_{i,j} = \underset{n}\max(\bold{\Bar{C}}_{i,n})\\ 0 &\text{otherwise,} \end{cases} &
    \quad
    \bold{A^{2}}_{i,j} &= \begin{cases} 1 &\text{if } \bold{\Bar{C}}_{j,i} = \underset{n}\max(\bold{\Bar{C}}_{j,n})\\ 0 &\text{otherwise,} \end{cases}
\end{aligned}
\end{equation}
}

\begin{equation}
    \bold{A}_{i,j} = \bold{A^{1}}_{i,j} \times \bold{A^{2}}_{i,j}
\end{equation}

The matrix $\bold{A}$ gives  the two final sets of matched points $\Bar{\boldsymbol{X}} \in \mathbb{R}^K$ and $\Bar{\boldsymbol{Y}} \in \mathbb{R}^K$ by re-indexing the original point clouds $\boldsymbol{X} \in \mathbb{R}^M$ and $\boldsymbol{Y} \in \mathbb{R}^N$ with the row and column indices of the non-zero values of the matrix $\bold{A}$. An example of a performed matching is depicted in Fig.~\ref{fig.matching}. 
Once the sets of matched points are built, different techniques can be used to determine the rigid transformation. A classical SVD to the cross-covariance matrix between the centred subsets $\Bar{\boldsymbol{X}}$ and $\Bar{\boldsymbol{Y}}$ is used in MDGAT~\citep{mdgat}, while DCP~\citep{Wang_2019_ICCV} suggested a differentiable and soft SVD where the weights of each point are determined by applying a \emph{Softmax} to the score matrix $\bold{C}$. An alternative method is to apply a RANSAC technique based on the matched features, as reported in~\citep{huang2021predator, li2022wsdesc}. In our method, we use RANSAC based on our predicted correspondences to reduce the computational cost. Moreover, instead of considering all the $K$ matched points, we only use the $K^c$ most relevant ones allowing us to filter the outliers before the first iteration such that the transformation is performed in 500 iterations maximum. 
%

\section{EXPERIMENTS}\label{sec.exp}
\subsection{Dataset and Parametrisation}
To assess RoCNet, we first use the synthetic ModelNet40 dataset~\citep{wang2019prnet} containing 40 CAD models of different objects. The database comprises 12,311 3D point clouds divided into 9,843 sets for training and 2,468 for testing. Each of these point clouds is scaled to fit inside a sphere of radius $r = 1$~m. For each source point cloud called $\boldsymbol{X}$, a target point cloud $\boldsymbol{Y}$ is created by applying a random rigid transformation with a rotation ranging from 0$^\circ$ to 45$^\circ$ around each axis and a translation ranging from 0~cm to 50~cm in each direction. Finally, a random permutation of the points is performed. In the end, a point cloud of 1024 points is generated for each object. In addition, to be able to evaluate the robustness of registration or pose estimation methods, the source point clouds are reduced by a range of points emulating partial occlusions. 

The following lists the different configurations used to assess our method in terms of accuracy, robustness and generalization: 
\begin{itemize}
\item {\emph{Clean data}: synthetic data in ideal conditions.}

\item \emph{Partial overlap}: To simulate partial occlusions, we sample the 768 nearest neighbors of a random point in both the source and target point clouds.
\item \emph{Partial overlap and noisy data}: Clipped Gaussian noise with a range of $[-0.05, 0.05]$, a mean of $\mu = 0$, and a variance of 0.01 is added to each point of the sub-sampled point clouds. 
%
%
\item \emph{Real data}: The proposed method has also been evaluated using the real-world Stanford Bunny dataset~\citep{turk1994zippered} which consists of 10 point clouds of $\approx$ 100k points obtained by a range scanner. 
\end{itemize}

\subsection{Implementation details} \label{Implementation_details}

RoCNet is implemented with PyTorch on a Nvidia Tesla V100-32G GPU and trained using Adam optimizer~{\citep{kingma2014adam}} for 30 epochs on clean data and for 80 epochs on noisy data with a learning rate of $10^{-4}$ in both cases. The number of predicted correspondences used as input in RANSAC is $K^c=256$. The parameters $d$ and $L$ are set to $96$ and $6$ respectively. For the normal estimation, the neighbours are collected under a radius $r=30$ cm and a maximum of $K_{nn} = 128$ points are used in the covariance matrix. 
%
\subsection{Metrics} \label{metriques}
To benchmark our RoCNet architecture against state-of-the-art methods, we opt for two metrics widely used in the literature: the Root Mean Square Error (RMSE) and the Mean Absolute Error (MAE). They are used to compute the difference in rotation and translation between the estimated transformation and the provided ground truth (GT) one. 
%
\subsection{Results}\label{sub.sec.results}
RoCNet is assessed in different configurations (i.e., favourable and unfavourable) and compared to the main DL-based methods of the state-of-the-art and with the traditional ICP approach. The recent work VRNet~\citep{zhang2022vrnet} has nicely summarised  the performance of most of the related methods reported in the literature. 
We use these performances as a baseline and add performances of recently published methods such as WsDesc~\citep{li2022wsdesc} and R-PointHop~\citep{r_pointhop}. Note that in all the tables, $\bold{R}$ is given in degrees, and $\bold{t}$  in meters, while the best results are highlighted in bold and the second ones are underlined.
%
\subsubsection{Model trained on clean data}
The first configuration consists of the evaluation of RoCNet when the model is trained on clean data and no occlusions. 
\small
\begin{table}[!h]
\caption{Performances of the models trained on  clean data and full overlapping point clouds.}

\begin{center}
\begin{adjustbox}{width=0.6\textwidth}
\begin{tabular}{l | l  l | l l} 
\hline 

\hline 

\hline
Method & RMSE($\bold{R}$) & MAE($\bold{R}$) & RMSE($\bold{t}$) & MAE($\bold{t}$)  \\ 
\hline
ICP'92      & 12.28  & 4.613      & 0.04774 & 0.00228 \\
PTLK'19       & 13.75  & 3.893      & 0.01990 & 0.00445 \\
DCP-V2'19    & 1.090  & 0.752     & 0.00172 & 0.00117 \\
PRNET'19     & 1.722  & 0.665  & 0.00637 & 0.00465 \\
R-PointHop'22 & 0.340  & 0.240      & \underline{0.00037} & 0.00029 \\ 
VRNet'22    & \underline{0.091} & \underline{0.012} & \textbf{0.00029} & \textbf{0.00005} \\
\hline
\textbf{Ours} & \textbf{0.082} & \textbf{0.011} & 0.00047 & \underline{0.00008} \\
\hline 

\hline 

\hline
\end{tabular}
\end{adjustbox}
\label{tab.all.clean.data}
\end{center}
\end{table}
\normalsize

Table~\ref{tab.all.clean.data} provides the results. Our method outperforms the other methods in rotation with an improvement of a dozen per cent for the RMSE and MAE \emph{versus} the second best method VRNet~\citep{zhang2022vrnet}. However, VRNet remains the best in translation although the difference is slight compared to RoCNet, specifically for MAE($\bold{t}$) where RoCNet is second. 
%
\subsubsection{Model trained on clean data and under partial occlusions} 
In this assessment, we evaluate the methods behaviour when only a part of points is shared by the two point clouds to be aligned. This simulates, for example, partial occlusions. From Table~\ref{tab.partial.clean.data}, it can be highlighted that RoCNet outperforms significantly the other methods in all the metrics. Overall, our method reduces the registration error by roughly half in comparison with the second-ranked methods, \textit{i.e.}, VRNet and R-PointHop.    
\small
\begin{table}[!h]
\caption{Performances of the models trained on clean data and partially overlapping point clouds.}

\begin{center}
\begin{adjustbox}{width=0.6\textwidth}

\begin{tabular}{l | l  l | l l}
\hline 

\hline 

\hline
Method & RMSE($\bold{R}$) & MAE($\bold{R}$) & RMSE($\bold{t}$) & MAE($\bold{t}$)  \\ 
\hline
ICP'92        & 33.683  & 25.045 & 0.293 & 0.2500 \\
PTLK'19         & 16.735  & 07.550 & 0.045 & 0.0250 \\
DCP-V2'19       & 06.709  & 04.448 & 0.027 & 0.0200 \\
PRNET'19       & 03.199  & 01.454 & 0.016 & 0.0100 \\
R-PointHop'22 & 01.660   & \underline{00.350} & 0.014 & \underline{0.0008} \\
VRNet'22        & \underline{00.982}  & 00.496 & \underline{0.006} & 0.0039 \\
WsDesc'22      & 01.187  & 00.975     & 0.008 & 0.0070 \\

\hline
\textbf{Ours} & \textbf{00.412}  & \textbf{00.133} & \textbf{0.002} & \textbf{0.0002} \\
\hline 

\hline 

\hline
\end{tabular}
\end{adjustbox}
\label{tab.partial.clean.data}
\end{center}
\end{table}
\normalsize
%
\subsubsection{Model trained on noisy data and under partial occlusions} 
The last configuration concerns the evaluation of the proposed method under partial overlap using noisy data. As can be seen in Table~\ref{tab.partial.noisy.data}, RoCNet outperforms the other methods on all metrics, both in rotation and translation. RoCNet allows significant enhancement of the registration error, ranging from two-thirds to one-quarter compared to the method ranked second, i.e., WsDesc and even more in comparison to VRNet. This can be explained by the robustness of RoCNet to partial occlusions or noise or both at the same time. 
\small
\begin{table}[!h]
\caption{Performances of the models trained on noisy and partially overlapping point clouds.}
\begin{center}
\begin{adjustbox}{width=0.6\textwidth}

\begin{tabular}{l | l  l | l l}
\hline 

\hline 

\hline
Method & RMSE($\bold{R}$) & MAE($\bold{R}$) & RMSE($\bold{t}$) & MAE($\bold{t}$)  \\ 
\hline
ICP'92& 33.067 & 25.564 & 0.294 & 0.250 \\
PTLK'19  & 19.939 & 9.076 & 0.057 & 0.032 \\
DCP-V2'19 & 06.883 & 4.534 & 0.028 & 0.021 \\
PRNET'19& 04.323  & 2.051 & 0.017 & 0.012 \\
VRNet'22& 03.615 & 1.637 & 0.010 & 0.006 \\
WsDesc'22 & \underline{03.500}  & \underline{0.759} & \underline{0.006} & \underline{0.004} \\
\hline
\textbf{Ours} & \textbf{01.810}  & \textbf{0.620} & \textbf{0.004} & \textbf{0.003} \\
\hline 

\hline 

\hline
\end{tabular}
\end{adjustbox}

\label{tab.partial.noisy.data}
\end{center}
\end{table}
\normalsize
%
\subsubsection{Model trained on ModelNet40 and tested on the Stanford Bunny dataset} 
To evaluate RoCNet on the Stanford Bunny dataset, a random rigid transformation is applied to the source point clouds and 2048 points out of the initial 100K points are used to estimate the rigid transformation. The results reported in Table~\ref{tab:bunny_metrics} show that RoCNet is also efficient on real-world unseen data  outperforming state-of-the-art methods in three out of four of the used metrics. 
\small
\begin{table}[!h]

\caption{Performances on the Bunny Stanford dataset for the models trained on ModelNet40}

\begin{center}
\begin{adjustbox}{width=0.6\textwidth}

\begin{tabular}{l | l  l | l l}
\hline

\hline 
 
\hline
Method & RMSE($\bold{R}$) & MAE($\bold{R}$) & RMSE($\bold{t}$) & MAE($\bold{t}$)  \\ 
\hline

ICP'92  & 13.32 & 10.72 &  0.0492 & 0.0242 \\
FGR'16 & 1.99 & 1.49  & 0.1993 & 0.1658 \\
DCP'19  & 6.44 & 4.78  & 0.0406 & 0.0374 \\
R-PointHop'22  & 1.49 & 1.09  & 0.0361 & \textbf{0.0269} \\
\hline
\textbf{Ours} &  \textbf{0.99} & \textbf{0.83}  & \textbf{0.0338} & 0.0288 \\

\hline

\hline 
 
\hline

\end{tabular}
\end{adjustbox}
\label{tab:bunny_metrics}
\end{center}
\end{table}
\normalsize

\begin{figure}[!h]
\includegraphics[width=\columnwidth]{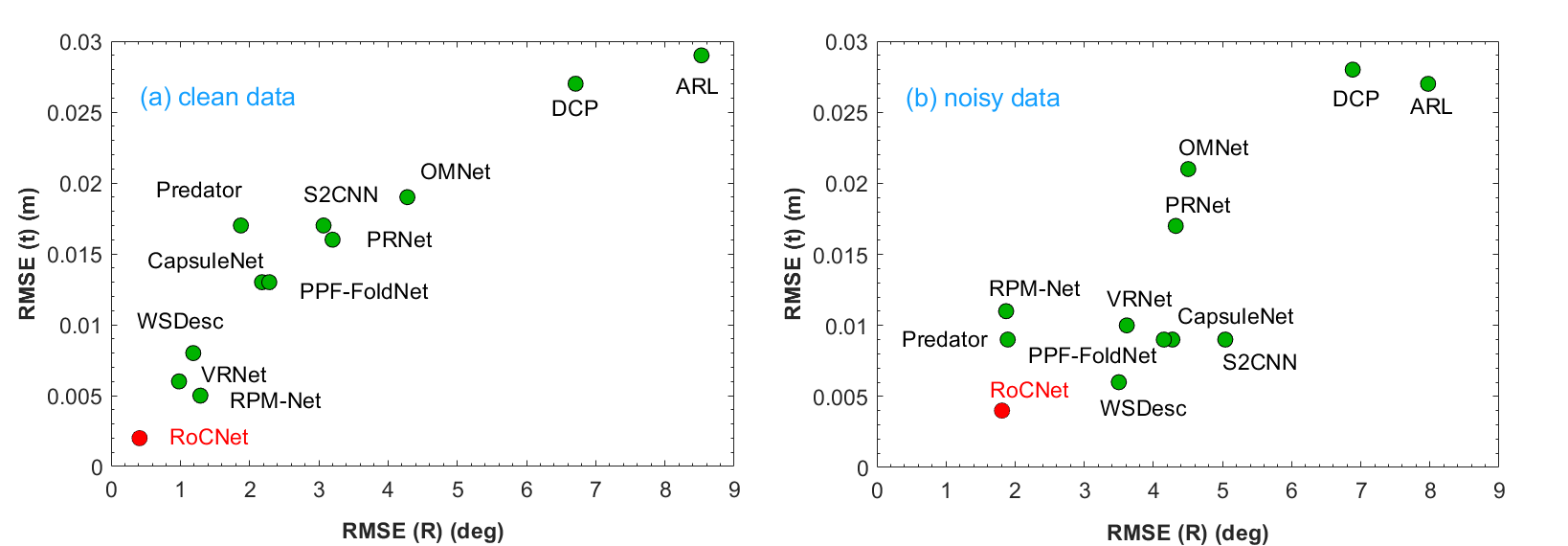}
\caption{Comparison of RoCNet performances against the eleven most relevant state-of-the-art methods. (a) in the case of clean data and (b) in the case of data with Gaussian noise, both with partial overlap. }
\label{fig.SoA}
\end{figure}

Finally, the performance of RoCNet and its ranking in the context of wider state-of-the-art, including the eleven best methods, is presented in Fig.~\ref{fig.SoA}. It can be highlighted that our method outperforms all the methods when considering simultaneously performances in rotation and translation, this both on clean data (Fig.~\ref{fig.SoA}(a)) and on noisy data (Fig.~\ref{fig.SoA}(b)). 


%

\begin{figure}[!h]
\centering
\setlength{\tabcolsep}{0.15em}
\renewcommand{\arraystretch}{0.8}

\begin{tabular}{ccccccccccccc}
{}
&
{\footnotesize{95\%}} &
{\footnotesize{80\%}} &
{\footnotesize{75\%}} &
{\footnotesize{65\%}} &
{\footnotesize{60\%}} &
{\footnotesize{55\%}} &
{\footnotesize{50\%}}
\\
{\rotatebox{90}{\footnotesize{~~~Inputs}}}
&
\includegraphics[width=0.13\columnwidth]{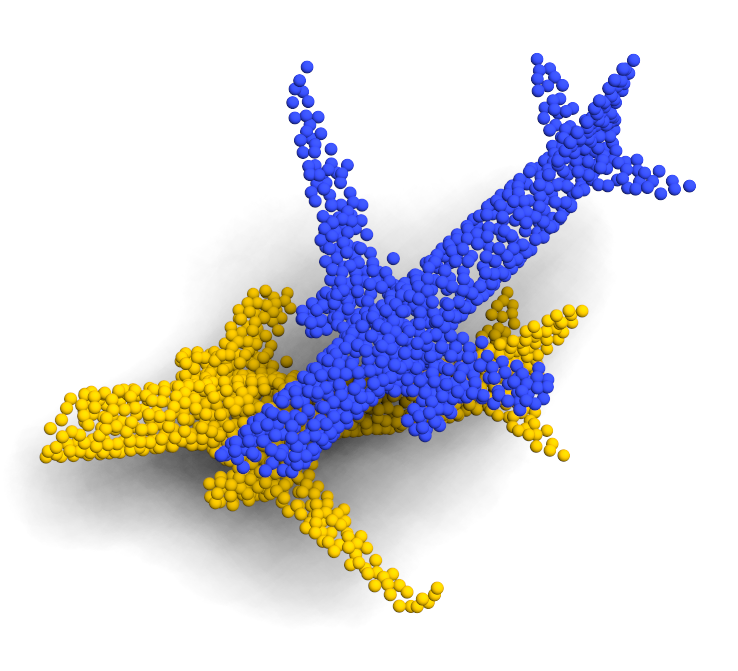}
&
\includegraphics[width=0.13\columnwidth]{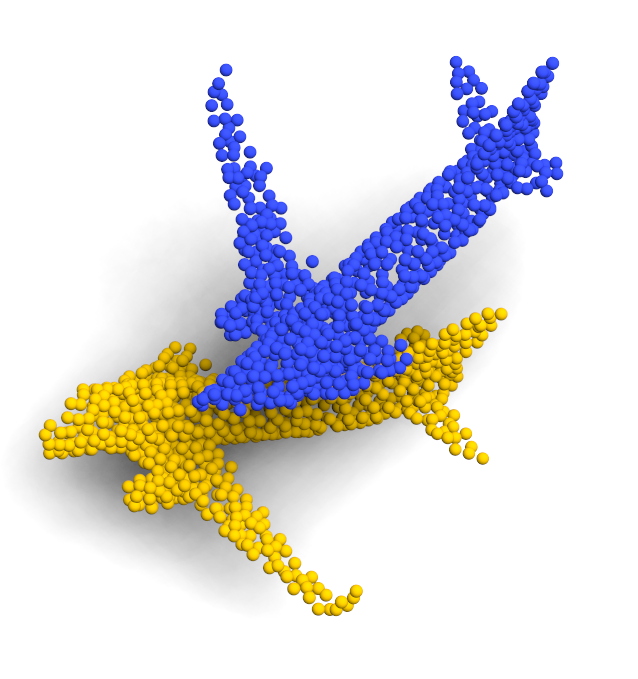}
&
\includegraphics[width=0.13\columnwidth]{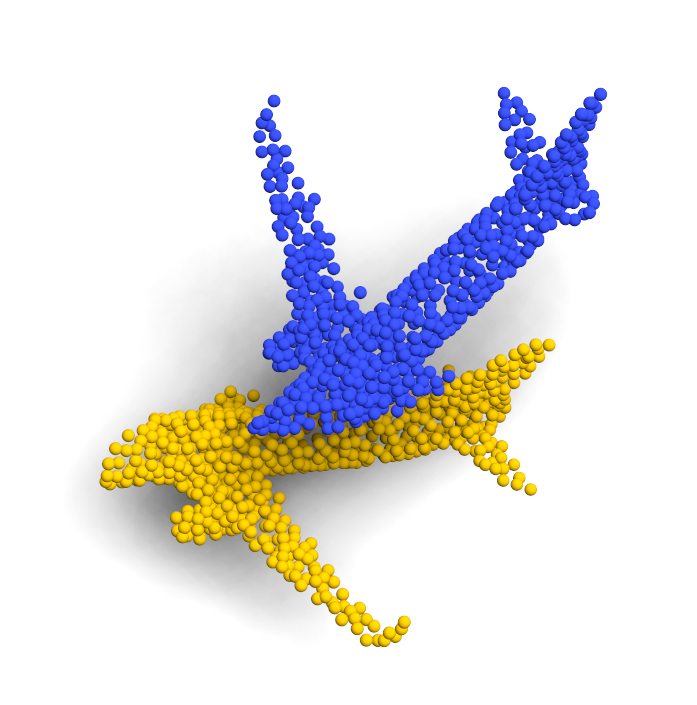}
&
\includegraphics[width=0.13\columnwidth]{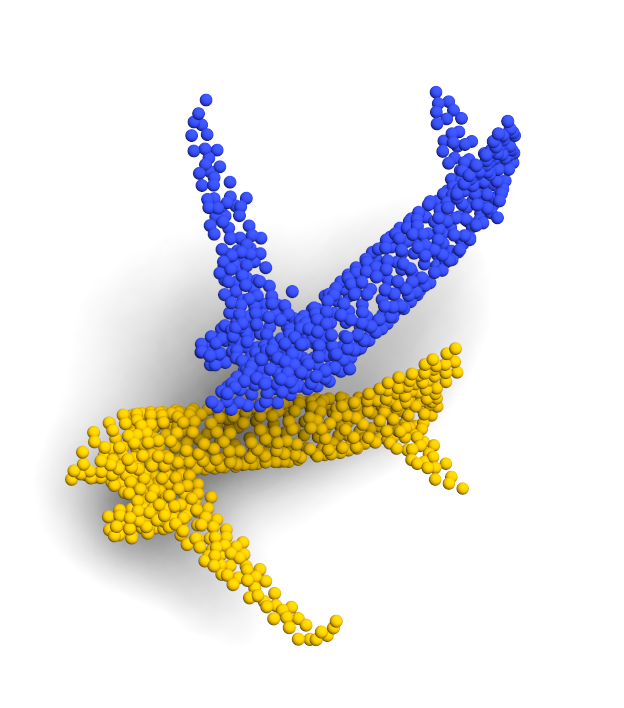}
&
\includegraphics[width=0.13\columnwidth]{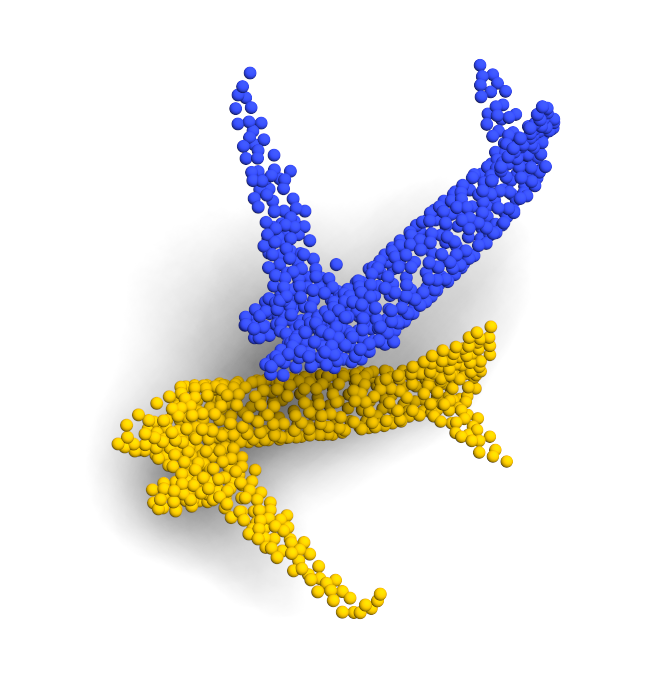}
&
\includegraphics[width=0.13\columnwidth]{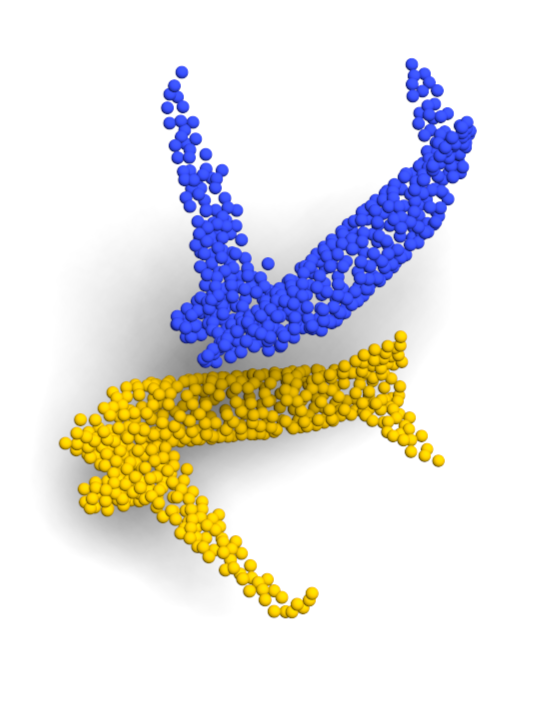}
&
\includegraphics[width=0.13\columnwidth]{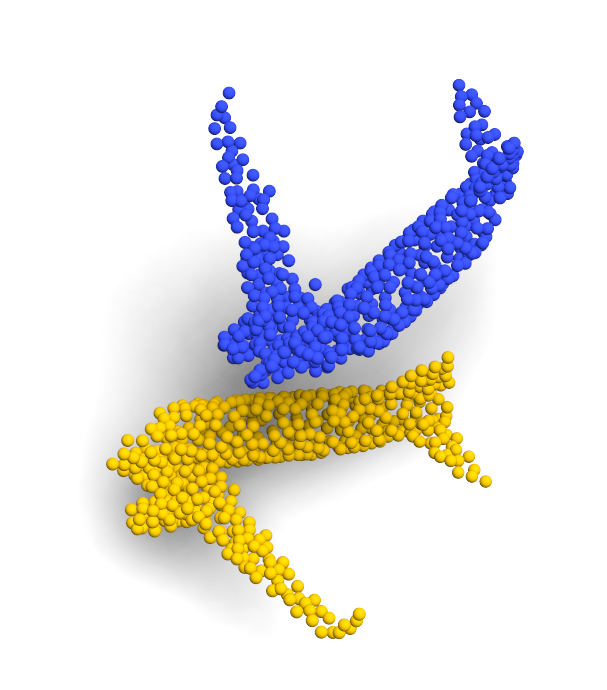}
\\[-2ex] 
{\rotatebox{90}{\footnotesize{Estimation}}}
&
\includegraphics[width=0.13\columnwidth]{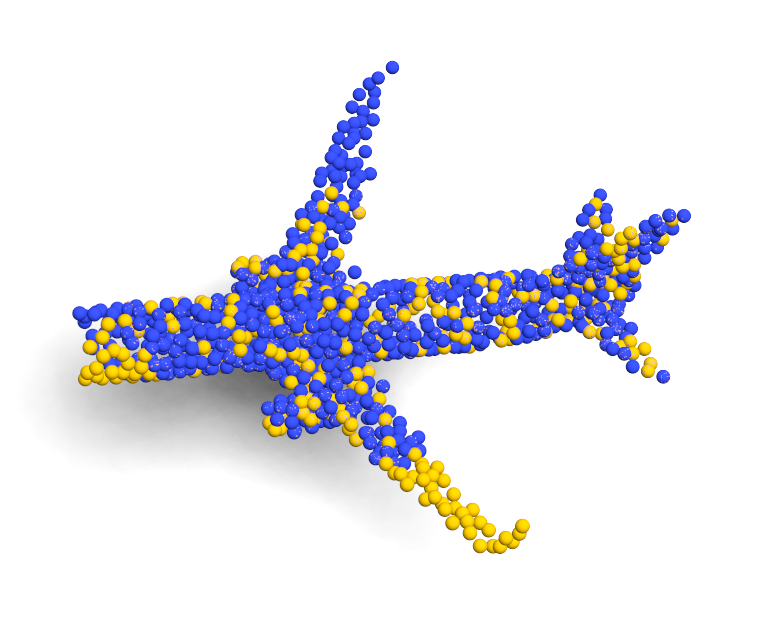}
&
\includegraphics[width=0.13\columnwidth]{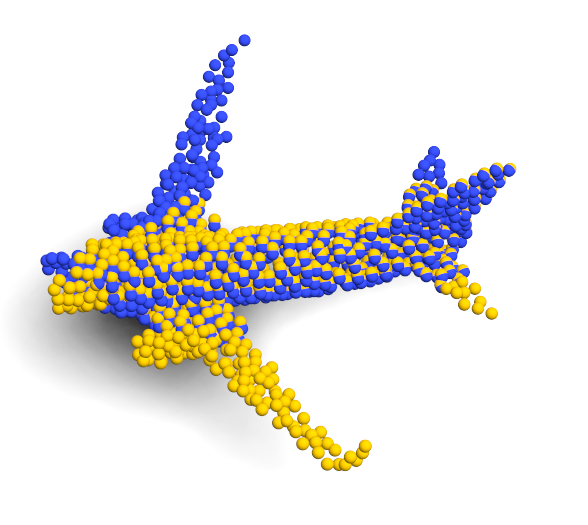}
&
\includegraphics[width=0.13\columnwidth]{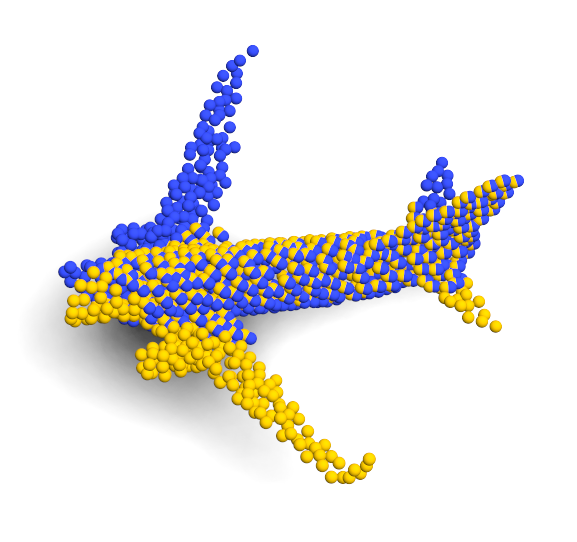}
&
\includegraphics[width=0.13\columnwidth]{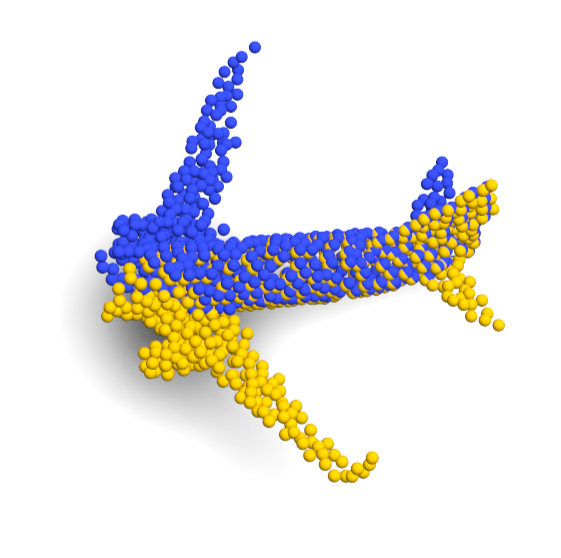}
&
\includegraphics[width=0.13\columnwidth]{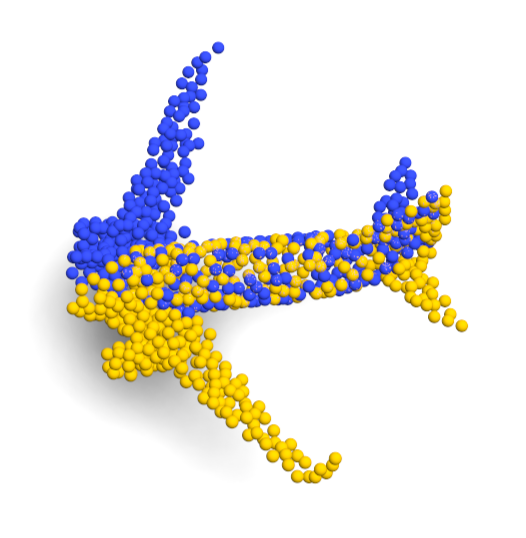}
&
\includegraphics[width=0.13\columnwidth]{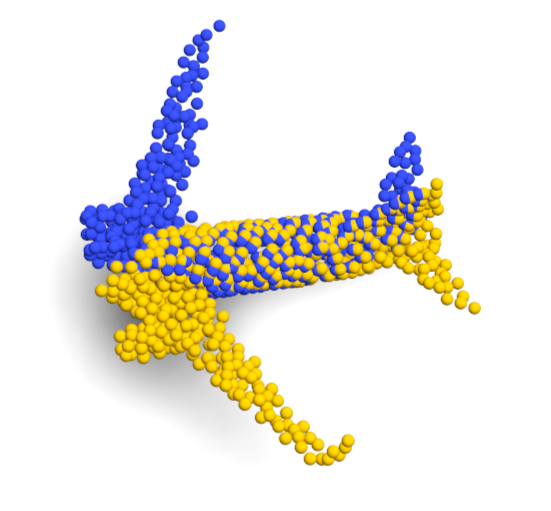}
&
\includegraphics[width=0.13\columnwidth]{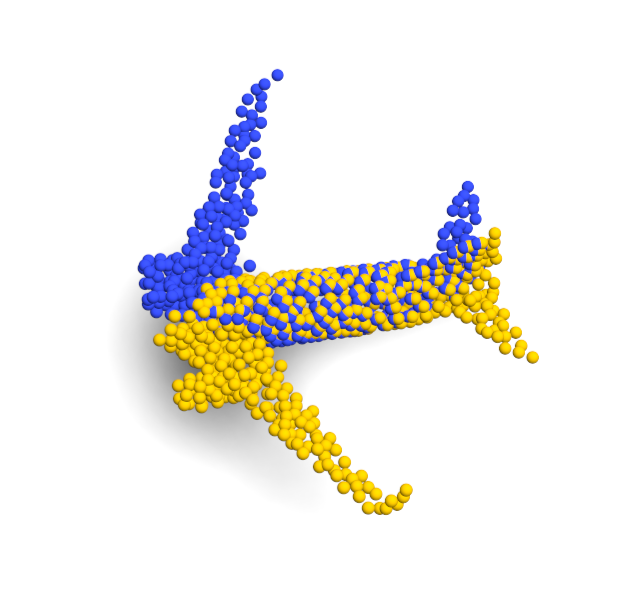}
\\[-2ex] 
{\rotatebox{90}{\footnotesize{~~GT}}}
&
\includegraphics[width=0.13\columnwidth]{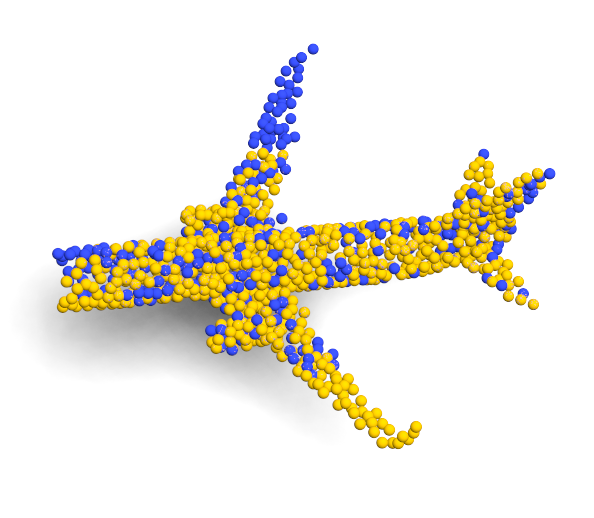} &
\includegraphics[width=0.13\columnwidth]{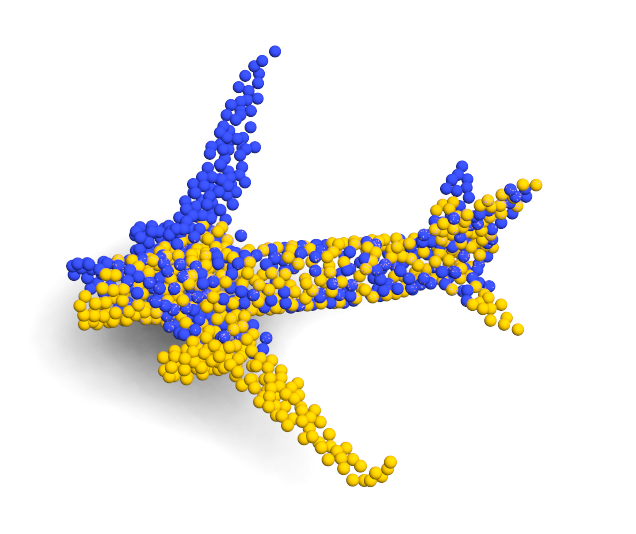} &
\includegraphics[width=0.13\columnwidth]{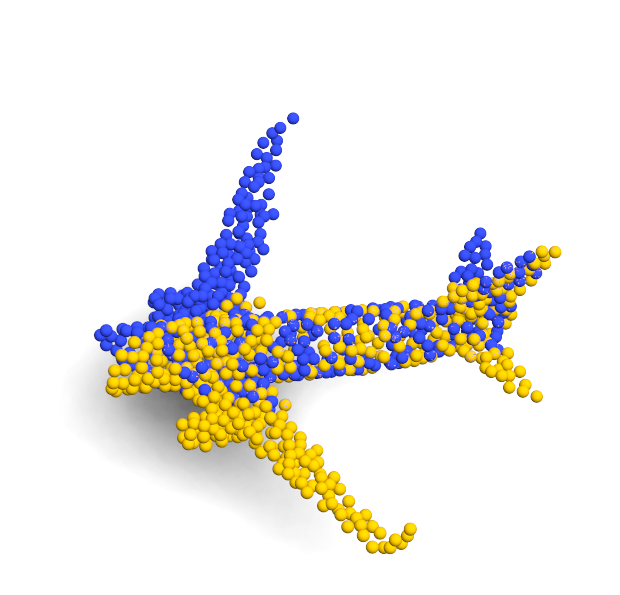} &
\includegraphics[width=0.13\columnwidth]{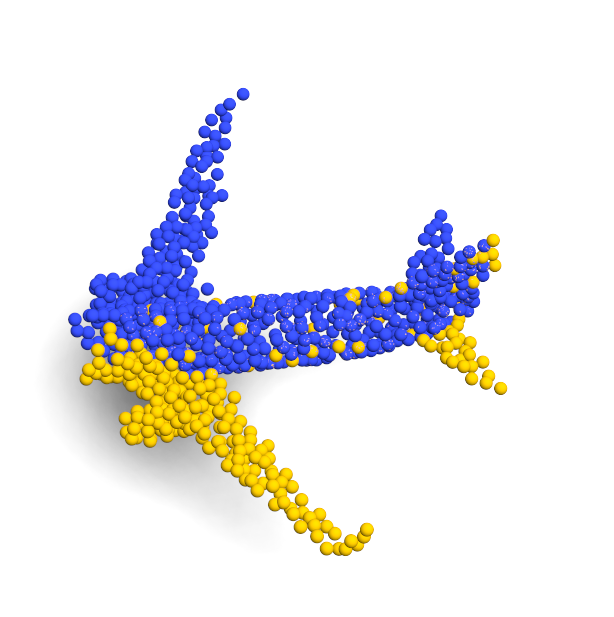} &
\includegraphics[width=0.13\columnwidth]{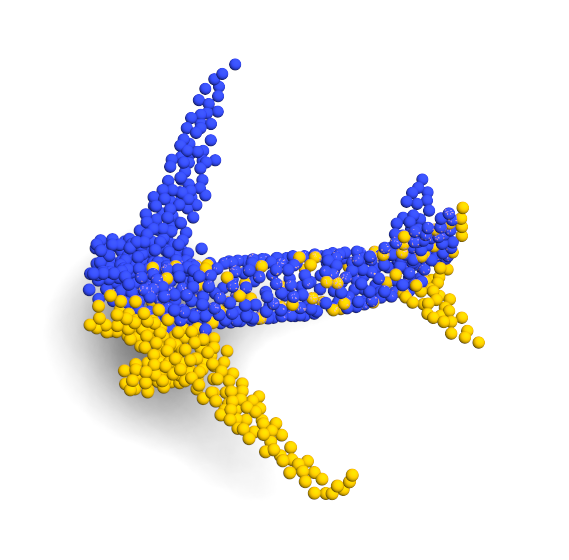} &
\includegraphics[width=0.13\columnwidth]{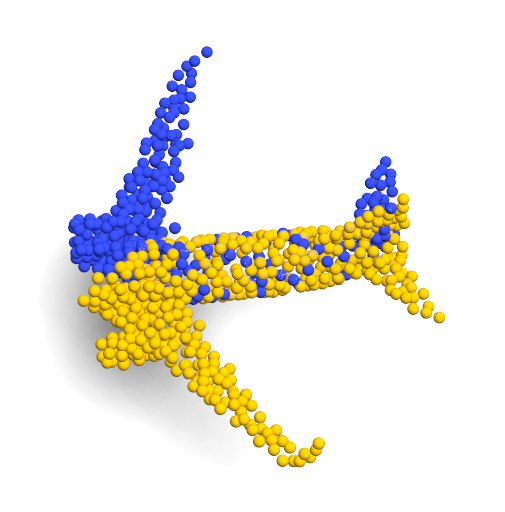} &
\includegraphics[width=0.13\columnwidth]{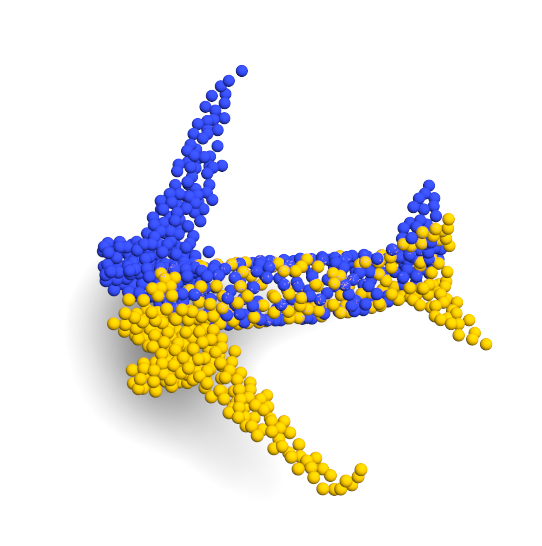}
\\[-1ex] 
\end{tabular}
\caption{Illustration of the robustness of RoCNet against partial occlusions (i.e., partial overall between the point clouds to be aligned).}
\label{fig.rate}
\end{figure}

Furthermore, to be able to assess visually the robustness ability of the proposed method, we perform different registrations by progressively decreasing (from 95$\%$ to 50$\%$) the rate of shared points between $\boldsymbol{X}$ and $\boldsymbol{Y}$. Figure~\ref{fig.rate} depicts the obtained results of one object. 
The first row shows the initial positions of source and target point clouds $\boldsymbol{X}$ and $\boldsymbol{Y}$, the second row shows the performed registrations and the third shows the ground truth ones. 
As can be seen, RoCNet can register data even with 50$\%$ points occluded without much difficulty. On the other hand, the method shows its limits for objects with perfect symmetry in low overlapping cases.

%

%
\section{ABLATION STUDY}\label{sec.ablation}
%
We conduct ablation studies on the three main blocks of the proposed architecture, i.e., the descriptor, the transformer, and the RANSAC-based estimation of the transformation.
%
\subsection{Descriptor ablation}
To study the impact of our descriptor which uses both normals and DGCNN as inputs on the transformer, we compared the performance of our architecture to that of MDGAT~\citep{mdgat} in case of point matching problem. For a proper comparison, both methods are trained in the same configuration and with the same number of epochs. Two types of data are used: 1) clean data and 2) noisy data, both with partial overlap. To achieve the comparison, we use the following metrics: Precision ($\mathbf{P}$), Accuracy ($\mathbf{A}$), Recall ($\mathbf{R}$) and F1-score ($\mathbf{F1}$).  Table~\ref{matching_compare} gives an insight into the ablation study of the descriptor. It can be underlined that our descriptor outperforms MDGAT one except for the Precision ($\mathbf{P}$) in the case of clean data with partial overlap. The difference is substantial, in favour of our descriptor when it concerns noisy data with partial overlap. 
\begin{table}
\centering
\caption{Performances assessment in a matching challenge with partial overlap matching (\textbf{P}: Precision, \textbf{A}: Accuracy, \textbf{R}: Recall, and \textbf{F1}: F1-score). }

\bigskip

\begin{adjustbox}{width=0.6\textwidth}
\begin{tabular}{l|cccc|cccc}
\hline 

\hline 

\hline

  & \multicolumn{4}{c|}{clean data} & \multicolumn{4}{c}{noisy data} \\ \cline{2-9} 
\multirow{-2}{*}{{ Method}} & \textbf{P} & \textbf{A} & \textbf{R} & \textbf{F1} & \textbf{P} & \textbf{A} & \textbf{R} & \textbf{F1} \\
\hline
\multicolumn{1}{c|}{MDGAT}  & \textbf{98.1} & 93.7 & 93.4 & 95.7 & 85.7 & 68.5 & 75.3 & 80.2 \\
\multicolumn{1}{c|}{Ours} & 98.0 &\textbf{97.2} &\textbf{97.6} &\textbf{97.8} & \textbf{85.8} & \textbf{85.9} & \textbf{85.5} & \textbf{85.7} \\

\hline 

\hline 

\hline

\end{tabular}
\end{adjustbox}
\label{matching_compare}
\end{table}

\small
\begin{table*}
\centering

\caption{ SVD \emph{versus} RANSAC for transformation estimation in case of full overlap, noisy data and partial overlap.}

\bigskip

\begin{adjustbox}{width=1\textwidth}

\begin{tabular}{l|ccc|ccc|ccc|ccc}
\hline 

\hline 

\hline
{ } & \multicolumn{3}{c|}{{ RMSE($\bold{R}$)}} & \multicolumn{3}{c|}{{ MAE($\bold{R}$)}} & \multicolumn{3}{c|}{{ RMSE($\bold{t}$)}} & \multicolumn{3}{c}{{ MAE($\bold{t}$)}} \\ \cline{2-13} 
\multirow{-2}{*}{{Method}} & {full} & {noisy} & {partial} & {full overlap} & {noisy} & {partial} & {full } & {noisy} & {partial} & {full } & {noisy} & {partial} \\
\hline
{SVD} & 3.94 & 1.94 & 4.21 & 0.124 & \textbf{0.484} & 0.168 & \textbf{0.0005} & \textbf{0.0017} & \textbf{0.001} & \textbf{0.0001} & \textbf{0.0018} & \textbf{0.0001} \\
{RANSAC} & \textbf{0.06} & \textbf{1.92} & \textbf{0.40} & \textbf{0.010} & 0.555 & \textbf{0.133} & \textbf{0.0005} & 0.0026 & 0.002 & \textbf{0.0001} & \textbf{0.0018} & 0.0002 \\
\hline 

\hline 

\hline
\end{tabular}
\label{tab.abla.ransac}
\end{adjustbox}
\end{table*}
\normalsize
%
\subsection{DGCNN and Transformer ablation} RoCNet is compared to other alternatives in which our descriptor and attention mechanism are changed to those proposed in~\citep{mdgat}. In view of evaluating the contribution of the proposed attention mechanism,  DGCNN is associated with a classical mechanism without the normals gradient embedding~\citep{sarlin20superglue}.  Table~\ref{tab.abla.dgcnn} reports the obtained results showing that RoCNet architecture is more relevant on three of the four used metrics emphasizing a significant contribution (about $10\%$) of the association of DGCNN and normals compared to a classical attention mechanism.
\small
\begin{table} 
\caption{Performances assessment in a matching challenge with clean data and partial overlap.}
\begin{center}
\begin{adjustbox}{width=0.6\textwidth}
    
\begin{tabular}{l| l |c c c c}
\hline 

\hline 

\hline
\textbf{Descriptor} & \textbf{{Attention}}& $\mathbf{P}$ & $\mathbf{A}$ & $\mathbf{R}$ & $\mathbf{F1}$ \\
\hline
MDGAT’s descriptor & \textbf{Ours} & 92.1 & 84.6 & 84.8 & 88.3 \\

\textbf{Ours} & MDGAT & 96.6 & \underline{96.3} & \underline{96.3} & 95.4\\
\textbf{Ours} & \textbf{Ours} w/o normals  & 79.8 & 80.2 & 80.5 & 80.1\\
\hline
 MDGAT’s descriptor & MDGAT & \textbf{98.1} & 93.7 & 93.4 &  \underline{95.7}  \\
\hline 
\textbf{Ours} & \textbf{Ours} & \underline{98.0} & \textbf{97.2 }& \textbf{97.6} & \textbf{97.8}  \\
\hline 

\hline 

\hline

\end{tabular}
\end{adjustbox}

\label{tab.abla.dgcnn}
\end{center}
\end{table}
\normalsize
%
\subsection{RANSAC ablation} 
The last ablation study consists of the comparison of the contribution of an SVD \emph{versus} RANSAC to estimate the rigid transformation when the matching is performed. Table~\ref{tab.abla.ransac} reports the performances of each alternative using: 1) clean data with full overlap (full), 2) noisy data with full overlap (noisy) and 3) clean data with partial overlap (partial). It can be seen that RANSAC approach  outperforms slightly the SVD one. 
%
\section{CONCLUSION}
%
This paper presents a new 3D point cloud geometric registration learning method. The proposed architecture is composed of three main blocks: 1) the newly designed descriptor which encodes the neighbourhood of each point and an attention mechanism that encodes the variations of the surface normals, 2) the matching method that estimates a matrix of correspondences using the Sinkhorn algorithm, and 3) the estimation of the rigid transformation using a RANSAC applied to the best matches from the correspondence matrix. The proposed architecture has been evaluated using the ModelNet40 dataset in different favourable and unfavourable configurations. It has been demonstrated that our method outperforms the related state-of-the-art algorithms, especially in unfavourable conditions, e.g., with noisy data and partial occlusions. 
{RoCNet has been evaluated on synthetic data augmented with noise and occlusion to simulate real ones and has proved its potential. One limitation is that DGCNN features combined with the proposed transformer leads to a huge memory solicitation and can not be used on large points clouds as in odometry.}  

{For future work, we plan to extend RoCNet by incorporating into the architecture an additional component for extracting a limited number of points (keypoints) to which the transformers can be effectively applied. This will allow the proposed method to be applied to dense real-world data, under the same conditions outlined in the paper, and we hope to achieve the same performance gain as  in this article.
}


\section*{Acknowledgments}
This work was supported by the French ANR program MARSurg (ANR-21-CE19-0026).


\bibliography{rocnet}  

\begin{thebibliography}{10}

\bibitem{besl1992method}
P.~J. Besl and N.~D. McKay, ``Method for registration of 3-d shapes,'' in {\em Sensor fusion IV: control paradigms and data structures}, vol.~1611, pp.~586--606, 1992.

\bibitem{wang2019prnet}
Y.~Wang and J.~M. Solomon, ``Prnet: Self-supervised learning for partial-to-partial registration,'' {\em Adv. Neural. Inf. Process. Syst.}, vol.~32, 2019.

\bibitem{Wang_2019_ICCV}
Y.~Wang and J.~M. Solomon, ``Deep closest point: Learning representations for point cloud registration,'' in {\em IEEE Int. Conf. on Comput. Vision}, 2019.

\bibitem{aoki2019pointnetlk}
Y.~Aoki, H.~Goforth, R.~A. Srivatsan, {\em et~al.}, ``Pointnetlk: Robust \& efficient point cloud registration using pointnet,'' in {\em IEEE/CVF Conf. Comput. Vision Pattern Recognit.}, pp.~7163--7172, 2019.

\bibitem{dgcnn}
Y.~Wang, Y.~Sun, {\em et~al.}, ``Dynamic graph cnn for learning on point clouds,'' {\em ACM Trans. on Grap.}, 2019.

\bibitem{qi2017pointnet++}
C.~R. Qi, L.~Yi, H.~Su, {\em et~al.}, ``Pointnet++: Deep hierarchical feature learning on point sets in a metric space,'' {\em Adv. Neural. Inf. Process. Syst.}, vol.~30, 2017.

\bibitem{soft_svd}
T.~Papadopoulo and M.~L. Lourakis, {\em Estimating the jacobian of the singular value decomposition: Theory and applications}.
\newblock PhD thesis, Inria, 2000.

\bibitem{turk1994zippered}
G.~Turk and M.~Levoy, ``Zippered polygon meshes from range images,'' in {\em Conf. on Comp. Grap. and Inter. Tech.}, pp.~311--318, 1994.

\bibitem{rusu2009fast}
R.~B. Rusu, N.~Blodow, {\em et~al.}, ``Fast point feature histograms (fpfh) for 3d registration,'' in {\em IEEE Int. Conf. on Rob. and Auto.}, pp.~3212--3217, 2009.

\bibitem{qi2016pointnet}
C.~R. Qi, H.~Su, {\em et~al.}, ``Pointnet: Deep learning on point sets for 3d classification and segmentation,'' in {\em Conf. Comput. Vision Pattern Recognit.}, pp.~652--660, 2017.

\bibitem{mdgat}
C.~Shi, X.~Chen, K.~Huang, {\em et~al.}, ``Keypoint matching for point cloud registration using multiplex dynamic graph attention networks,'' {\em IEEE Rob. and Auto. Let.}, vol.~6, pp.~8221--8228, 2021.

\bibitem{r_pointhop}
P.~Kadam, M.~Zhang, S.~Liu, {\em et~al.}, ``R-pointhop: A green, accurate, and unsupervised point cloud registration method,'' {\em IEEE Trans. on Ima. Process.}, vol.~31, pp.~2710--2725, 2022.

\bibitem{li2022wsdesc}
L.~Li, H.~Fu, and M.~Ovsjanikov, ``Wsdesc: Weakly supervised 3d local descriptor learning for point cloud registration,'' {\em IEEE Trans. Vis. Comput. Graph.}, 2022.

\bibitem{huang2021predator}
S.~Huang, Z.~Gojcic, M.~Usvyatsov, {\em et~al.}, ``Predator: Registration of 3d point clouds with low overlap,'' in {\em IEEE/CVF Conf. Comput. Vision Pattern Recognit.}, pp.~4267--4276, 2021.

\bibitem{yew2018-3dfeatnet}
Z.~J. Yew and G.~H. Lee, ``3dfeat-net: Weakly supervised local 3d features for point cloud registration,'' in {\em Euro. Confe. on Comput. Vision}, 2018.

\bibitem{triplet_loss}
M.~Khoury, Q.-Y. Zhou, and V.~Koltun, ``Learning compact geometric features,'' in {\em Proceedings of the IEEE international conference on computer vision}, pp.~153--161, 2017.

\bibitem{bai2020d3feat}
X.~Bai, Z.~Luo, L.~Zhou, {\em et~al.}, ``D3feat: Joint learning of dense detection and description of 3d local features,'' in {\em IEEE/CVF Conf. Comput. Vision Pattern Recognit.}, pp.~6359--6367, 2020.

\bibitem{roufosse2019unsupervised}
J.-M. Roufosse, A.~Sharma, and M.~Ovsjanikov, ``Unsupervised deep learning for structured shape matching,'' in {\em Proceedings of the IEEE/CVF International Conference on Computer Vision}, pp.~1617--1627, 2019.

\bibitem{ovsjanikov2012functional}
M.~Ovsjanikov, M.~Ben-Chen, J.~Solomon, A.~Butscher, and L.~Guibas, ``Functional maps: a flexible representation of maps between shapes,'' {\em ACM Trans. on Grap.}, vol.~31, pp.~1--11, 2012.

\bibitem{geotransformer}
Z.~Qin, H.~Yu, C.~Wang, {\em et~al.}, ``Geometric transformer for fast and robust point cloud registration,'' in {\em IEEE/CV Conf. Comput. Vision Pattern Recognit.}, pp.~11143--11152, 2022.

\bibitem{zhou2021scanet}
R.~Zhou, X.~Li, and W.~Jiang, ``Scanet: a spatial and channel attention based network for partial-to-partial point cloud registration,'' {\em Pattern Recognition Letters}, vol.~151, pp.~120--126, 2021.

\bibitem{vaswani2017attention}
A.~Vaswani, N.~Shazeer, {\em et~al.}, ``Attention is all you need,'' {\em Adv. Neural. Inf. Process. Syst.}, vol.~30, 2017.

\bibitem{sarlin20superglue}
P.-E. Sarlin, D.~DeTone, {\em et~al.}, ``{SuperGlue}: Learning feature matching with graph neural networks,'' in {\em Conf. Comput. Vision Pattern Recognit.}, 2020.

\bibitem{zhao2021point}
H.~Zhao, L.~Jiang, {\em et~al.}, ``Point transformer,'' in {\em IEEE/CVF Int. Conf. Comput. Vision}, pp.~259--268, 2021.

\bibitem{sinkhorn1967concerning}
R.~Sinkhorn and P.~Knopp, ``Concerning nonnegative matrices and doubly stochastic matrices,'' {\em Pacific J. of Math.}, vol.~21, pp.~343--348, 1967.

\bibitem{kingma2014adam}
D.~P. Kingma and J.~Ba, ``Adam: A method for stochastic optimization,'' {\em arXiv preprint arXiv:1412.6980}, 2014.

\bibitem{zhang2022vrnet}
Z.~Zhang, J.~Sun, Y.~Dai, {\em et~al.}, ``Vrnet: Learning the rectified virtual corresponding points for 3d point cloud registration,'' {\em IEEE Trans. on Cir. and Sys. for Video Tech.}, vol.~32, pp.~4997--5010, 2022.

\end{thebibliography}






\end{document}